\documentclass[11pt]{article}

\usepackage[margin=1in]{geometry}
\usepackage{amsmath}
\usepackage{amssymb}
\usepackage{caption}
\usepackage{subcaption}
\usepackage{graphicx}
\usepackage{tabularx}
\usepackage{booktabs}
\usepackage{hyperref}
\usepackage{natbib}
\usepackage{xcolor}
\usepackage{enumitem}
\usepackage{adjustbox}

\title{On Benchmarking Human-like Intelligence in Machines}

\author{%
Lance Ying\textsuperscript{1,2,*},
Katherine M.\ Collins\textsuperscript{2,5},
Lionel Wong\textsuperscript{7},
Ilia Sucholutsky\textsuperscript{4},
Ryan Liu\textsuperscript{5},\\[0.3em]
Adrian Weller\textsuperscript{3},
Tianmin Shu\textsuperscript{6},
Thomas L.\ Griffiths\textsuperscript{5},
Joshua B.\ Tenenbaum\textsuperscript{2}\\[1.0em]
{\small\itshape \textsuperscript{1}Harvard University \quad
\textsuperscript{2}Massachusetts Institute of Technology}\\[0.2em]
{\small\itshape \textsuperscript{3}University of Cambridge \quad
\textsuperscript{4}New York University \quad
\textsuperscript{5}Princeton University}\\[0.2em]
{\small\itshape \textsuperscript{6}Johns Hopkins University \quad
\textsuperscript{7}Stanford University}%
}

\date{}

\begin{document}

\maketitle

\renewcommand{\thefootnote}{\fnsymbol{footnote}}
\footnotetext[1]{Corresponding author. E-mail: lanceying@g.harvard.edu}
\renewcommand{\thefootnote}{\arabic{footnote}}
\setcounter{footnote}{0}

\begin{abstract}
Recent advances in Artificial Intelligence (AI) have yielded powerful computational models that, by learning from vast amounts of human-generated data, are increasingly posited as approximate models of human cognition. However, we argue that many current evaluation paradigms for AI are insufficient for assessing human-like cognitive capabilities in these models. We identify a set of key shortcomings: a lack of human-validated labels, inadequate representation of human response variability and uncertainty, and reliance on simplified and ecologically invalid tasks. We support our claims by conducting a human evaluation study on nine existing AI benchmarks, suggesting major limitations in task and label designs. To address these limitations, we propose five concrete recommendations for future AI evaluation efforts that will enable more rigorous and meaningful understanding of human-like cognitive capacities in AI models.
\end{abstract}

\vspace{0.5em}

\section{Introduction}
\label{intro}

From the earliest days of cognitive science and artificial intelligence (AI), the vision of creating machines that think and act like humans has captured the imagination of researchers and the public alike \citep{turing1950, lake2017building, cave2023imagining, weizenbaum1966eliza, anderson1990cognitive}. This pursuit is driven not only by scientific curiosity (to better understand and model human intelligence) but also by the potential of human-like AI to reshape our world, through the ways that we engage with our work and with each other.

Recent years have witnessed a surge in claims that AI systems have achieved human-like behavior on various tasks. However, the relevance of these results for determining whether AI systems act in a way that is human-``like'' is challenged by the limitations of existing evaluation benchmarks.

In this perspective paper, we argue that \textbf{current evaluation tools for AI systems are often insufficient for assessing human-like cognitive capabilities and future evaluation benchmarks can draw lessons from rigorous cognitive science research}. Specifically, we highlight a few shortcomings: the too-frequent absence of human validation in dataset labeling, inadequate representation of human variability in collected human data, and over-reliance on simplified tasks that lack ecological validity and fail to reflect the complexity of real-world scenarios. We support these claims with a human evaluation study on $9$ well-known AI benchmark tasks, showcasing potential limitations along these three axes. To address these critical gaps, we propose five concrete recommendations for the development of future benchmarks, derived from best practices in cognitive modeling. We believe these recommendations will pave the way for more rigorous and meaningful evaluations of human-like AI for both cognitive scientists and AI researchers. Importantly, our recommendations are specifically targeted at benchmarks whose goal is to assess whether AI systems exhibit human-like cognitive capabilities, i.e., benchmarks that evaluate whether AI systems reason, perceive, or make decisions in ways that are qualitatively similar to humans. We acknowledge that benchmarks designed to measure domain-specific capabilities where the goal is superhuman performance (e.g., mathematical theorem proving, protein structure prediction) or benchmarks that evaluate abstract reasoning at the level of the best human experts (e.g., ARC-AGI) may have different evaluation goals, and our recommendations may not apply to those settings in the same way. We close with open questions and challenges of implementing these recommendations.

\begin{table}[ht!]
\centering
\begin{tabular}{p{4cm}p{8cm}}
\hline
Task & Description\\
\hline
\multicolumn{2}{l}{\textbf{BIG-Bench} \citep{srivastava2022beyond}} \\
Fantasy reasoning & Reason about scenarios that violate the ordinary rules of the world \\[1mm]
Social IQA & Reason about typical social situations. \\[1mm]
Moral permissibility & Reason about morally permissible actions in scenarios \\[1mm]
Simple ethical questions & Give perspectives on a set of hypothetical, consequential, political, and social questions.\\
Social support & Distinguish supportive and unsupportive language uses. \\[1mm]
Irony identification & Determine whether a text is meant to be ironic or not. \\[1mm]
Dark humor detection & Detect whether a particular piece of text is intended to be humorous (in a dark way) or not
\\[1mm]
\hline
\multicolumn{2}{l}{\textbf{ToMBench} \citep{chen2024tombench}} \\
Ambiguous story task & Reason and answer questions about ambiguous social situations \\[1mm]
\hline
\multicolumn{2}{l}{\textbf{BigToM} \citep{gandhi2024understanding}} \\
Theory of Mind reasoning & Answer questions about agents' beliefs and actions \\
\hline
\end{tabular}
\caption{Benchmark tasks used in our experiment to evaluate human response distributions and levels of agreement.}
\label{tab:benchmarks}
\end{table}

\section{Building and Evaluating Human-like AI}
\label{human-like AI}

There has been a long history of interest in building and evaluating human-like intelligence in machines. But what do we mean by human-like intelligence? There is no universal definition; indeed, whether behavioral similarity alone is sufficient for attributing intelligence has been debated since at least \citet{block1981psychologism}, who argued that a system could produce human-like outputs without any genuine cognitive processing. We argue that part of this disagreement arises because ``human-likeness'' can be assessed at different levels of analysis. A useful framework for thinking about this comes from \citet{marr1982vision}, who proposed three levels at which any information-processing system can be understood: the \textit{computational level} (what problem is being solved and why), the \textit{algorithmic/representational level} (what representations and processes are used to solve it), and the \textit{implementational level} (how the system is physically realized). A machine could be human-like at any one of these levels without necessarily being human-like at the others. For instance, a system might arrive at human-like judgments through entirely non-human internal representations.

Much of the current work in AI evaluation has focused on the \textit{behavioral level}, assessing whether AI systems produce outputs that are similar to those of humans on cognitive tasks. This corresponds most closely to Marr's computational level: characterizing the input-output mapping that the system computes, without committing to claims about how that mapping is internally achieved. In this paper, we adopt this behavioral perspective. Specifically, we define human-likeness at the behavioral level as the capacity of an AI system to produce judgments and behaviors that are qualitatively similar to those of humans across a range of cognitive tasks, capturing not only which response the system gives, but also its patterns of errors, uncertainties, and the distributional variability observed across human populations. On the other hand, human-likeness at representational level, often called \textit{representational alignment} (whether a model's internal representations are structured similarly to those of humans), is also an important and growing area of research \citep{sucholutsky2023getting, muttenthaler2024aligning}.


It is also important to distinguish \textbf{human-like} from \textbf{human-level} intelligence. ``Human-level'' typically refers to achieving quantitative performance parity with humans, matching or exceeding human accuracy on a task. ``Human-like,'' by contrast, refers to qualitative similarity in cognitive patterns: not just getting the same answers right, but exhibiting similar patterns of uncertainty, similar distributions of responses across a population, and similar error profiles. A system could achieve human-level performance on a mathematical reasoning task by solving problems correctly using methods entirely unlike human cognition; such a system would be human-level but not necessarily human-like. Conversely, a system that makes errors in human-like ways and shows graded uncertainty similar to humans may be human-like even if its overall accuracy is lower.

But why may we aim for human-like AI? The pursuit of human-like AI is motivated by both scientific curiosity and practical considerations. From the earliest days of AI, scholars have sought to understand, model, and attempt to replicate the intricacies of human cognition and intelligence \citep{rosenblatt1958perceptron,10.7551/mitpress/4943.003.0128,minsky1988society, mitchell2024turing} and use these cognitively-informed models for practical applications. Building human-like AI offers a powerful lens through which to explore fundamental questions about the philosophy of mind, the nature of human cognition, and the underlying mechanisms driving complex human behavior. This quest not only pushes the boundaries of computer science but also promises to deepen our understanding of human intelligence.

Creating AI systems that exhibit human-like thinking and behaviors offers several potential advantages for applications. Human-like AI can think and act instead of humans in many scenarios while ensuring safety and reliability:

\begin{itemize}

    \item Effective Human-AI Interaction: Humans have developed complex social cognitive skills for effective collaboration, which involves simulating other agents' mental states and future actions \citep{bandura2001social, gallese2007before}. AI systems that adhere to human-like patterns of reasoning and behavior can enable human users to easily construct accurate mental models of the AI partner and better simulate and predict the AI partner's future actions \citep{collins2024building}. This leads to more effective collaboration and coordination between human users and AI agents \citep{carroll2019utility, ho2022cognitive, zhi2024pragmatic}. Additionally, interacting with agents that behave predictably and understandably can reduce cognitive load \citep{dragan2013legibility, fisac2020generating}. We don't have to expend as much mental effort trying to decipher unfamiliar or unexpected behaviors.

    \item Better simulated agents: AI systems with human-like cognitive capabilities are valuable tools for building simulations of people. This has many benefits, including improving communication \citep{liu2023improving, shaikh2024rehearsal}, generating feedback on pilot studies, and even potentially automating human participant responses in social sciences \citep{ashokkumar2024predicting, park2024generative, demszky2023using} or Human Computer Interaction \citep{hamalainen2023evaluating}. Prior work has also explored the use of LLMs for product testing \citep{brand2023using} and substituting human subjects in software engineering \citep{gerosa2024can}.

\end{itemize}

That said, it remains unclear exactly which domains would most critically benefit from human-like AI. We argue that there are two complementary perspectives. First, from a scientific standpoint, rigorously comparing machine intelligence to human cognition across the full range of cognitive tasks, including error patterns, biases, and uncertainty, is valuable for advancing our understanding of both human and machine intelligence, regardless of practical application. Second, from a practical standpoint, benchmarking human-likeness is especially important for subjective tasks that lack a normative standard of correctness, such as judgments of moral permissibility, social appropriateness, or humor. In these domains, evaluating AI against a fixed ground-truth label is fundamentally misguided, and human-like distributional evaluation becomes not just informative but necessary. This contrasts with tasks that do have a normative standard of correctness, such as mathematical theorem proving, program synthesis, or protein structure prediction, where reproducing human error patterns may even be counterproductive.

Given the scientific and practical benefits of building human-like AI, evaluating whether models genuinely exhibit human-like traits and behaviors is a critical and growing area of interest for both cognitive science and AI. Historically, many AI benchmarks have been developed to measure model performance on tasks requiring human-like cognitive capabilities (e.g.\ language understanding, common sense reasoning, etc.). These AI benchmarks typically rely on simple accuracy metrics, comparing the model's output against a single ground-truth label, often produced by curators or human annotators. This approach provides a straightforward measure of performance which facilitates easy comparison across different models. In this paper, we will critically examine these common evaluation practices and discuss how future AI benchmarks can be significantly improved by integrating insights and methods from cognitive science research.
\section{Benchmark Selection and Evaluation}

To motivate our recommendations, we collected human data on $9$ commonly used AI benchmarks. We selected $7$ benchmarks from BIG-Bench \citep{srivastava2022beyond} under the common-sense reasoning category and two Theory-of-Mind reasoning benchmarks, BigToM \citep{gandhi2024understanding} and ToMBench \citep{chen2024tombench}. The benchmarks are described in Table~\ref{tab:benchmarks}. We chose these benchmarks because they target inherently subjective cognitive tasks, such as moral reasoning, social cognition, and humor detection, as well as Theory of Mind reasoning, that we believe would benefit most from rigorous practices drawn from cognitive science, including the collection of human-validated labels, measurement of response uncertainty, and evaluation against population-level distributions. Many focus on language understanding and social cognition, which are particularly pertinent for human-AI interaction. All $9$ benchmarks have a single ground truth label for each stimulus.

We randomly sampled $30$ stimuli from each benchmark and recruited $117$ participants from Prolific, yielding $2{,}685$ annotated trials. The human study was approved by an IRB. Each participant was paid at a rate of 15 USD per hour and completed a consent form before the annotation tasks. We used the same answer options provided by the benchmarks, but instead of using a multiple choice question we asked participants to drag a slider on a scale from $0$--$100$ (e.g.\ $0$ = strongly disagree, $100$ = strongly agree) for each answer option. Additionally, after each slider response, participants were asked to report their confidence in their judgment on a separate scale from $0$ (not at all confident) to $100$ (extremely confident). This two-measure design allows us to disentangle graded judgments (e.g., feature intensity or degree of belief) from epistemic uncertainty (how confident a participant is in that judgment), addressing a potential confound inherent in slider-only elicitation methods.

We highlight some aggregate statistics and diagnostic examples in the section below to support our arguments. More detailed analysis and examples can be found in the Appendix.

\begin{figure}[t!]
    \centering
    \includegraphics[width=0.85\textwidth]{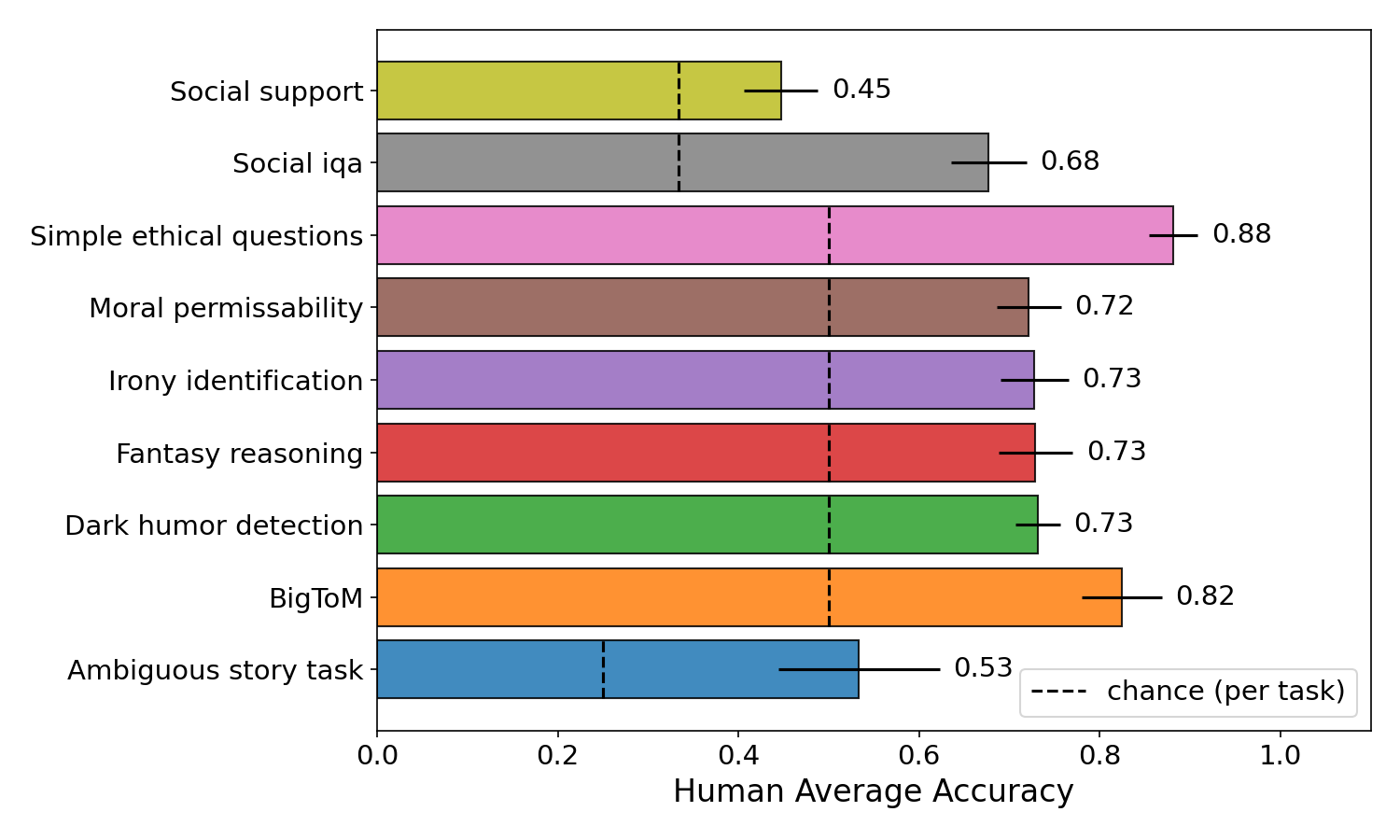}
    \caption{Human average accuracy per task. Error bars indicate 95\% confidence intervals. Dashed black ticks indicate per-task chance levels.}
    \label{fig:accuracy-by-task}
\end{figure}

\section{Pitfalls and Recommendations for Benchmarking Human-like AI}
In this section, we present recommendations for evaluating ``human-like'' AI. More recently, a growing body of work has offered cognitive-science-inspired recommendations for AI evaluation that overlap with and complement our proposals. \citet{frank2023baby} argued for careful, incremental evaluation of language model capacities using methods from developmental psychology. \citet{ivanova2025evaluate} outlined guidelines for evaluating the cognitive abilities of LLMs, emphasizing construct validity and controlled experimental design. \citet{shiffrin2023probing} discussed the challenges of using psychological methods to probe AI models and the interpretive pitfalls that can arise. From the comparative cognition perspective, \citet{voudouris2024comparative} and \citet{rane2024principles} have advocated for applying principles from animal cognition research to AI evaluation, emphasizing the importance of carefully designed behavioral assays that avoid anthropocentric assumptions. Other work has highlighted specific methodological challenges in human-machine comparison: \citet{firestone2020performance} drew an important distinction between \textit{performance} and \textit{competence} in human-machine comparisons, arguing that observed task performance may not reflect underlying competence due to confounding factors in how tasks are administered to humans versus machines. Relatedly, \citet{hu2024auxiliary} showed that auxiliary task demands can mask model capabilities, \citet{lampinen2024can} demonstrated the importance of carefully matched comparison conditions between models and humans, and \citet{hu2023prompting} established that prompting methods are not equivalent to direct probability measurements when evaluating language models.

Our work builds on and extends these contributions. Here, we focus particularly on how insights from decades of computational modeling can inform how we approach AI evaluation. The recommendations we propose here derive from years of development and debate in cognitive science to determine best practices for designing tasks, richly comparing models to human judgments, and sharpening hypotheses about what aspects of human behavior a computational model is intended to capture in the first place: all cornerstones, we argue, of what it means to make theoretically rich, replicable, and measured claims about the sense in which a given model is and is not comparable to human behavior. We urge developers of AI evaluation benchmarks to engage with and capitalize on this history.

\begin{figure}[t!]
    \centering
    \includegraphics[width=\textwidth]{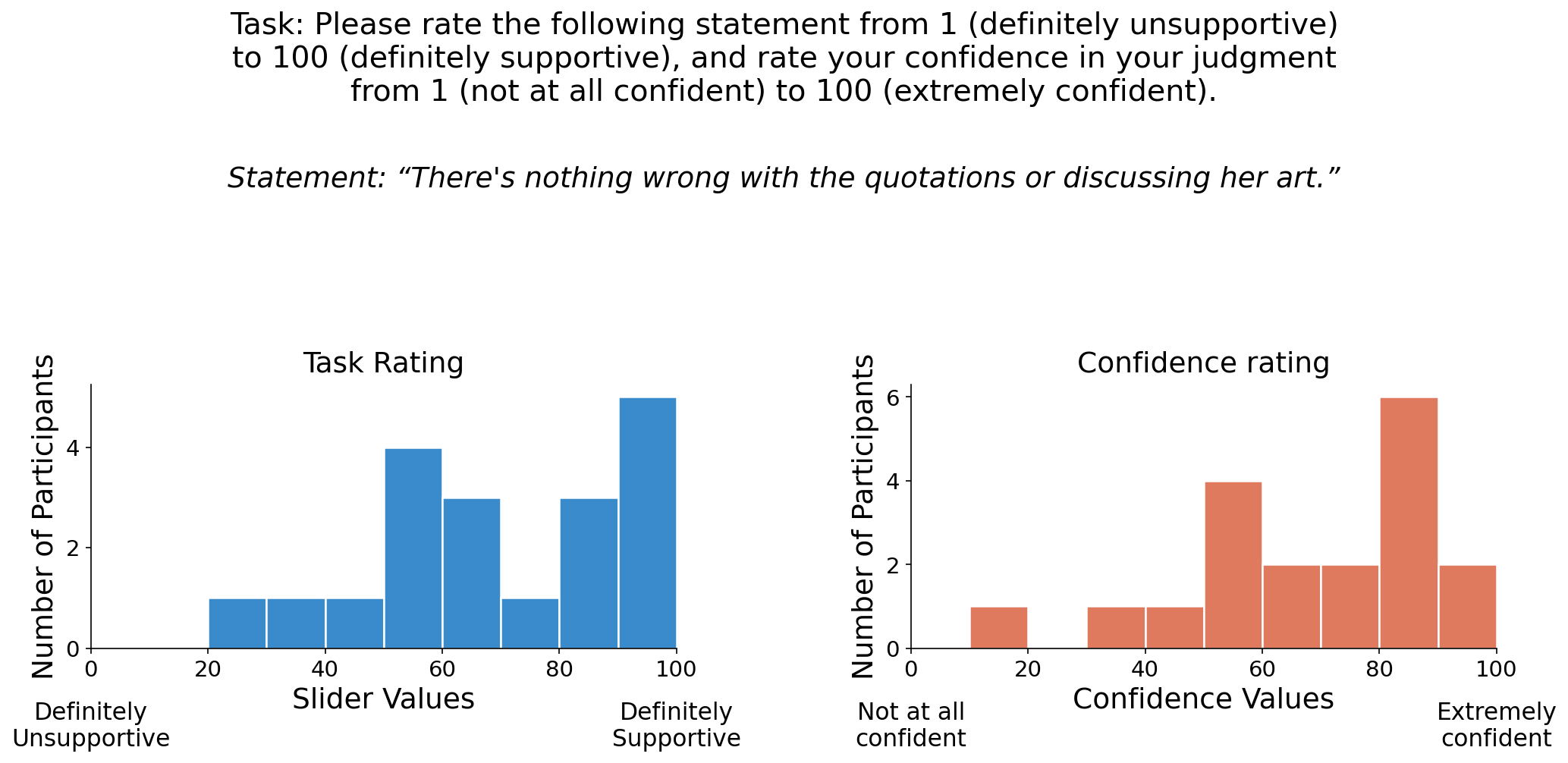}
    \caption{Distribution of participants' task ratings (left) and confidence ratings (right) on one stimulus from the Social support classification task. Participants are asked to rate how supportive is the statement ``There's nothing wrong with the quotations or discussing her art.'' The ground-truth label is ``unsupportive,'' yet most participants showcase graded judgment in the level of supportness, with the majority of the ratings fall between 50 to 100. The confidence histogram shows that people also report varying degrees on uncertainty.}
    \label{fig:social_support}
\end{figure}

\subsection{Recommendation 1: Measure `human-like AI' against actual humans, and collect robust, replicable sample sizes of human data}
\label{human-data}
A surprising number of ``cognitively-inspired'' benchmark suites and AI evaluations claim to measure human-like AI performance without any human data at all. Rather, tasks derived or sometimes loosely adapted from psychological assays are used to directly evaluate computational model performance, often with ground-truth notions of what it means to ``solve'' a task (for instance, to identify whether a model can label mental states in simple ``false-belief'' tasks derived from cognitive theory-of-mind experiments \citep{wimmer1983beliefs}). Our first and perhaps most fundamental recommendation is that the \textbf{ground-truth labels for measuring whether AI is human-like should be response data collected from humans themselves.}

Using actual human behavior as the ``gold'' labels for AI benchmarks, we propose, is important for many structural aspects that have been well documented in cognitive science. First, many AI benchmarks seek to evaluate inherently subjective concepts, such as whether an act is morally permissible, where a single, objectively correct answer (or even any set of ``correct answers'') may not exist. Rather, computational models of subjective behavior like moral reasoning, have long sought to characterize distributions of human judgments, including to account for known variation across populations, social groups, and cultures \citep{graham2009liberals,graham2016cultural}, while also seeking to explain how these differences arise \citep{levine2020logic}.

Second, even on tasks that appear to have a single objective ``gold label'' based on external measures, measuring human behavior may still reveal important variation and disagreement, sometimes with high confidence, that is nonetheless revealing of the internal computations by which humans process particular inputs. The famous visual illusion involving \href{https://en.wikipedia.org/wiki/The_dress}{The Dress}, for instance, illustrates people's strongly diverging judgments even given a measurable external label, the true color of the dress. These divergent judgments on this single stimulus reveal important, measurable, and modelable facets of human visual processing \citep{lafer2015striking}. Our own data demonstrates analogous patterns of structured disagreement across higher-level cognitive tasks (see Figure~\ref{fig:accuracy-by-task} and Figure~\ref{fig:distribution-label}). More generally, building systems that are truly human-like or that can well-model human-like behavior requires also modeling human error patterns and uncertainty. Computational cognitive modelers do not shy away from human errors, but rather lean into them; such errors can help reveal structure in what we do or do not know. Understanding whether a machine is human-like therefore ought to examine such error patterns from the ``true'' state of the world.

In our analysis of a suite of common AI evaluation benchmarks that had previously been annotated with only a single ``correct'' answer, we found high levels of disagreement between human judgments and benchmark labels. As shown in Figure~\ref{fig:accuracy-by-task}, human accuracy, measured as the percentage of trials where a participant's response agreed with the benchmark's ground-truth label (see Appendix for computation details), varies substantially across tasks, with an average of $69.7\%$ across all nine benchmarks. Notably, Social support ($45\%$, chance $= 50\%$) falls below chance, and Moral permissibility ($72\%$), Irony identification ($73\%$), Fantasy reasoning ($73\%$), Dark humor detection ($73\%$), and Social IQA ($68\%$, chance $= 33\%$) all show substantial disagreement between human annotators and benchmark labels. Only Simple ethical questions ($88\%$) approaches high agreement. These results underscore a fundamental concern: if we aim to evaluate whether AI systems are human-like, using labels that do not reflect how most humans actually respond introduces a misleading standard of comparison. Consider the specific example in Figure~\ref{fig:social_support}: participants are asked to rate whether the statement ``There's nothing wrong with the quotations or discussing her art'' is supportive. We find that people report graded levels of supportiveness and express varying degrees of uncertainty in their judgments. On average, participants rate the statement as moderately supportive (mean $= 69.4$, SD $= 21.6$ on a 0--100 scale, where 100 = definitely supportive), while the ground truth label in the dataset is ``unsupportive.'' We show more such examples in the Appendix.

Taken together, our re-annotation of these benchmarks, with real humans, suggests that there are serious concerns as to the validity of some published ground-truth labels for benchmarking ``human-likeness.'' This finding is consistent with broader evidence that pervasive label errors in benchmark test sets can destabilize machine learning evaluations more generally \citep{northcutt2021pervasive}.

\subsection[Recommendation 2: Evaluate models of human populations against population-level distributions of human judgments]{Recommendation 2: Evaluate models of human populations against\\ population-level distributions of human judgments}
\label{human-data-dist}
Our second recommendation builds more specifically on the inter-annotator variation we discuss above: for many AI models, particularly machine learning models explicitly trained on large distributions of human-generated data, we propose that model evaluations should explicitly collect, analyze, and use population-level distributions of human responses as the ``gold'' soft labels for evaluating model performance. A fundamental distinction for computational cognitive and psychological models is clarifying which populations of humans one seeks to model, and at what level one seeks to model them, distinguishing, for instance, between a granular model of the algorithms, strategies, and errors that a single human might make across related stimuli on a single domain, with the overall pattern of responses we can expect to find across many subjects. Because many AI models are trained on population-level human data using objectives designed to measure population-level responses, and are often intended for deployment across populations, we argue that it is crucial to collect and evaluate performance explicitly on how well models capture the structure and variation of behavior across sets of human subjects.

Nearly all facets of human cognition are influenced by a complex set of individual differences and cultural factors. These include differences in underlying cognitive abilities or resources like working memory or attention \citep{boogert2018measuring}; differences in prior experiences, preferences and goals, which can influence how individuals predict unknowns given limited evidence or choose among a set of options and actions \citep{ongchoco2024new}; and cultural variation in values, expectations, and experiences that systematically influence priors or decision making strategies \citep{henrich2010weirdest}.

Many existing benchmarks collect human annotations but rely on majority voting to collapse the human responses to a single ``ground-truth'' label, effectively discarding valuable information about the range and distribution of human judgments. This may disproportionally lead models to align with the majority view, even if there are important subpopulations that are otherwise underrepresented \citep{gordon2022jury}. Additional pitfalls of such information loss in label construction have been raised in the context of image classification systems wherein the labels used to train models were often taken to be the label with the majority vote; several works identified that training and evaluating such models on distributions over annotator uncertainty (``soft labels'') revealed and guarded against otherwise fragility in such model predictions \citep{peterson2019human, sucholutsky2023informativeness, collins2023humanMixup, Uma_Fornaciari_Hovy_Paun_Plank_Poesio_2020}. These works also highlight the potential benefit of then training on labels that better capture the richness of human beliefs for enhanced generalization and robustness. As a concrete example, \citet{peterson2019human} introduced CIFAR10H, a dataset providing full distributions of human labels for CIFAR-10 test images, and showed that classifiers trained on these soft labels generalize better to out-of-distribution test sets and exhibit improved robustness against adversarial attacks compared to those trained on majority-vote hard labels. We advocate for the consideration of distributions over human data in the context of AI evaluation more broadly.

Researchers in AI Alignment, specifically ``pluralistic alignment'', have advocated for similar recommendations \citep{kirk2024prism, sorensen2024roadmap} but more restricted to alignment to a distribution of values and preferences in decision-making. Relatedly, there is substantial work on calibration in machine learning, i.e., how well a model's answer uncertainty matches its accuracy, which is relevant when considering whether model uncertainty should match the human distribution \citep{geng2024survey}. Distributional evaluation and calibration assessment serve complementary goals: the former asks whether a model's response patterns resemble those of human populations, while the latter asks whether a model's expressed confidence is well-calibrated. In our paper, we argue modeling distributions over annotators should \textbf{extend beyond value-laden tasks to encompass the full range of cognitive tasks where human variability is meaningful}, such as tasks involving language understanding (where pragmatic interpretation varies across individuals), and social cognition (where inferences about others' mental states are inherently uncertain).

\paragraph{Designing and evaluating population-level metrics.} Once we collect the distribution of human data, how may we evaluate AI models? As in cognitive modeling, where researchers often deploy a range of evaluation measures on collected data and conduct analyses on subgroups within populations of participants, we recommend being clear and seeking explicitly to measure the following:

\begin{itemize}
    \item Report metrics used to compare distributions of samples from models (with comparable numbers of samples from the model versus samples from a population of participants) to distributions of human judgments, such as measures on probability distributions (e.g., KL divergence or Wasserstein distances). These metrics can ensure that models do not simply report narrow means, with little of the expected distributional diversity shown across populations as a whole.

    \item  Explain structure within a given distribution of answers. For instance, if distributions have distinct modes, can the model interpretably and consistently explain how these modes arise, or how modes are correlated across related questions?

    \item Measure how the model represents individual patterns of answers and explain individual differences across the population, for instance, to what degree can it capture conditional patterns based on personal traits (e.g.\ how a pluralist would answer a moral value judgment query versus a utilitarian)? Evaluating conditional distributions can help further focus which parts of a population are well-modeled, and which may be more divergent. Relatedly, a model conditioned on a given individual or subgroup should be internally consistent like that individual, since errors on cognitive tasks tend to be more correlated within a subject than across subjects. On this view, the population distribution is recovered by marginalizing over such conditioned agents, rather than emerging from a single, self-inconsistent set of samples.
\end{itemize}

\subsection{Recommendation 3: Evaluate model \textit{gradedness} and \textit{uncertainty} against gradedness in individual human judgments}

Just as different people may come to different conclusions about any given task, any single person may be uncertain about what decision they want to make or what plan they want to take. Decades of cognitive science research has shown that graded beliefs and uncertainties are an essential part of human cognition, driving nuanced human perception, reasoning and behaviors \citep{tversky1974judgment, chater2006probabilistic, griffiths2024bayesian}. We encourage benchmark builders to consider eliciting, maintaining, and measuring not just judgment over hard labels with multiple choice questions but graded judgments from individual annotators using soft labels. The collection and consideration of soft labels for capturing graded judgments from humans has been standard practice for cognitive modeling and has more recently been advocated for in the context of computer vision \citep{sucholutsky2023getting, uma2022scaling, schmarje2022one}, human-AI interaction \citep{steyvers2022bayesian, collins2023human}, and the elicitation of knowledge from experts more broadly \citep{uncertainJudgments, expertElicitation}. These judgements can also be collected and compared over hierarchical representations \citep{muttenthaler2024aligning}.

\begin{figure}[t!]
    \centering
    \includegraphics[width=\textwidth]{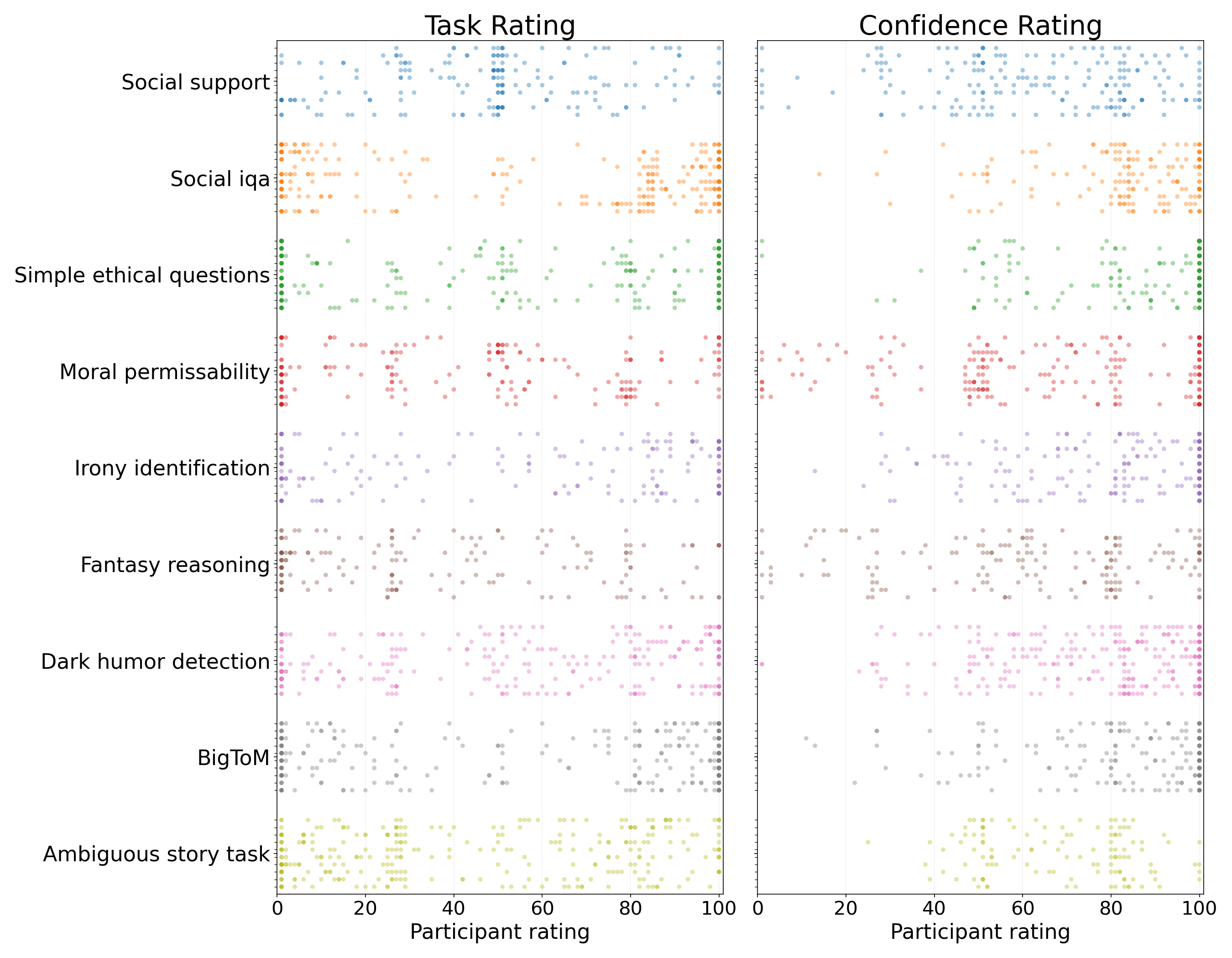}
    \caption{Distributions of participants' task ratings (left) and confidence ratings (right) across all 9 benchmarks. Each row represents a stimulus in a benchmark. Each dot represents a response from a participant on the stimulus. Task ratings show that participants hold graded beliefs poorly captured by binary ground-truth labels. Confidence ratings show that participants show varying degrees of uncertainty in the tasks, where tasks such as social support, moral permissibility, fantasy reasoning frequently receive low confidence ratings.}
    \label{fig:distribution-label}
\end{figure}

Discrete multiple-choice questions that require an annotator to select only one choice are typically too coarse for such measures. Figure~\ref{fig:distribution-label} shows the distribution of participants' slider ratings (left panel) alongside their confidence ratings (right panel) across all 9 benchmarks. Despite the fact that the underlying benchmark labels map onto extreme values of 0 or 100, a large fraction of human ratings fall in intermediate ranges, demonstrating that participants hold graded beliefs that are poorly captured by discrete labels. Importantly, the separate confidence ratings reveal that these graded responses are not simply artifacts of participant uncertainty: participants often report high confidence in their intermediate judgments, indicating that the gradedness reflects genuine degrees of belief or feature intensity rather than epistemic confusion. This pattern is pervasive across tasks.

We call on AI benchmarks to consider collecting and assessing soft labels from annotators to measure their graded judgments for the following reasons. First, graded judgments better reflect the nuances of real-world scenarios. Real-world decision-making rarely involves absolute, binary choices. Consider emotions, which vary in intensity, or moral judgments, where two wrong actions might warrant different levels of reprimand. Graded responses allow benchmarks to capture these crucial distinctions and nuances and can in turn be used to train models for better generalization to new situations \citep{peterson2019human, NEURIPS2023_9febda1c, muttenthalerhuman}.

Second, soft labels capture the inherent uncertainty prevalent in many tasks. A binary choice often fails to represent the full spectrum of human beliefs and judgment. Individuals may lean towards one option while acknowledging some doubt. This uncertainty is fundamental to real-world reasoning and decision-making. Quantifying uncertainty allows for flexible planning, adaptive strategies, and appropriate risk assessment, essential skills for robust AI systems. While some might argue that large samples with hard labels can approximate uncertainty, this approach hinges on the assumption of independent and identically distributed (i.i.d.) samples. However, this assumption often does not hold in many real-world cases due to individual and group-level variations. Consider the example of \href{https://en.wikipedia.org/wiki/The_dress}{The Dress}: averaging judgments across all samples would show high uncertainty between the two color labels, yet each individual is quite adamant about what they see. We also observe rich distributional patterns in our own data collection on cognitive tasks, for instance, in some tasks on Irony identification and Fantasy reasoning, participants cluster at both extremes of the scale with high confidence, suggesting that structured individual disagreement, not mere noise, underlies the variation.

To deeply understand whether a model is human-like, we urge \textbf{finer-grained consideration of the rich, structured beliefs that any single annotator may have.} Researchers may fear perceived ``messiness'' of collecting human uncertainty. An oft heard retort to the collection of uncertainty is that people are ``miscalibrated'' in their uncertainty. Decades of research in cognitive science, however, have designed studies to examine people's probabilistic judgments in order to study and model human cognition \citep{keren1991calibration,tenenbaum1998bayesian, chater2006probabilistic,windschitl1996measuring, uncertainJudgments, griffiths2024bayesian}. We encourage designers of AI benchmarks to engage with such literature and lean into these uncertainties in humans' judgments in order to assess models' human-like behaviors.

\subsection{Recommendation 4: Situate tasks with respect to meta-reviews of existing cognitive theory}

To draw generalizable conclusions about an AI model's cognitive capacity, tasks should be carefully designed to measure whether the model's cognitive capabilities are human-like \citep{hernandez2017evaluation}. To do so, benchmarks should begin with what we term a \textit{meta-theory} of the target cognitive construct. By meta-theory, we mean an explicit specification of the construct that defines what it is (and what it is not), decomposes it into subcomponents, and identifies the observable behavioral signatures that should arise under controlled manipulations, such as ablations, difficulty variations, resource constraints, or adversarial conditions. This framing draws on the classic literature on construct validity in psychometrics \citep{cronbach1955construct, messick1995validity}: a benchmark should be justified as a measurement instrument with an explicit argument for why its tasks validly measure the intended construct, rather than serving as a collection of intuitively related tasks \citep{bean2025measuring}. The meta-theory then guides the construction of the benchmark, ensuring that tasks effectively probe the specific cognitive capacities of interest and provide meaningful insights into to what extent AI possesses these mental constructs in a human-like way.

One common pitfall is the use of impoverished theory in guiding AI benchmark creation. For example, many benchmarks have been created to evaluate a machine's Theory of Mind (ToM), which refers to the human ability to make inferences about other agents' mental states. ToM benchmarks for AI commonly or exclusively use the Sally-Anne test (a.k.a.\ false-belief test) (e.g.\ \citealt{le2019revisiting}), which has traditionally been used in developmental psychology for evaluating the timing of children's developing Theory of Mind. The results from these evaluations have led to claims such as ToM having emerged in LLMs \citep{kosinski2024evaluating, gandhi2024understanding}. However, ToM embodies a wide range of subcomponents beyond those assessed by the Sally-Anne test. In a comprehensive review, \citet{beaudoin2020systematic} identified 220 ToM tasks and measures previously used by psychological studies. Other authors have also questioned the validity and effectiveness of the Sally-Anne test in assessing children's ToM \citep{bloom2000two}. By exclusively focusing on false-belief tasks, many studies on evaluating AI models' ToM reflect a poor understanding of the meta-theory of ToM as construed in cognitive psychology. A construct-valid evaluation of ToM would specify its subcomponents (e.g., false belief attribution, intention inference, desire reasoning, knowledge tracking), identify which behavioral signatures each subcomponent predicts under controlled conditions, and design tasks that probe these signatures while controlling for confounds. This is especially important because the target construct differs between developmental assays and model evaluation: Sally-Anne tasks were designed to detect a developmental milestone in children, not to provide a comprehensive measure of ToM competence. When applied to language models, they also risk conflating ToM with linguistic or pragmatic skills, as \citet{hu2025reevaluating} demonstrate, apparent model competence on these tasks often reflects language understanding rather than genuine mental state reasoning. Instead, benchmarking intelligent systems should \textbf{start from a meta-theory of the cognitive construct and design tasks grounded in the cognitive theory}, including a comprehensive survey of its subdomains, taxonomies, and measures. More broadly, if a construct is assumed to have a particular factor structure, as has been established for domains like self-regulation \citep{eisenberg2019uncovering}, then evaluations should verify that this structure holds for the models being tested, rather than assuming that human-derived constructs map directly onto model behavior. Whether human-derived meta-theories transfer to AI systems at all is itself an open question: in some cases, the constructs may not hang together in the same way, and validating the mapping from a human-derived meta-theory to a model's behavioral variation across tasks is an important prerequisite to any strong claim of ``human-likeness.''

Another common pitfall is the naive use and adaptation of psychological tests in evaluating AI models. Passing a few psychological tests is insufficient to claim certain cognitive capacities exist in machines. Again take the Sally-Anne test as an example. Although it may be effective in measuring children's ToM, tests as such are insufficient for evaluating AI's ToM because these well-known test items are likely present in the web-scraped training corpora of large language models, creating data contamination concerns that undermine the validity of such evaluations \citep{jacovi2023stop}. Unlike human children who encounter false-belief scenarios naturally during development, AI models may have been exposed to both the test items and their expected answers during training. Therefore, blindly taking psychological scales and applying them to AI benchmarks to claim an AI is human-like can result in misleading conclusions and the results will be unlikely to generalize to richer tasks in the real world. Instead, we encourage AI benchmark creators to use psychological theories as a guide and psychological tests as inspirations for designing tasks for evaluating AI's cognitive capacity, but the tasks should be richer, more grounded, and more complex. Research in Cognitive Science in the past decades have introduced many rich and interactive paradigms for studying and evaluating models' social cognition (e.g.\ \citealt{baker2017rational,jara2020naive,ying2023inferring}), which were used to extract sophisticated and graded reasoning patterns from humans. In the next section, we discuss some concrete recommendations for designing such tasks.

\subsection{Recommendation 5: Design Cognitively Rich Tasks that Elicit Naturalistic Behavior}
\label{task}

Benchmark tasks should elicit the kind of rich, structured cognitive behavior that humans display in the real world, rather than being constrained to simple, unambiguous question-answer formats. Many existing benchmarks focus on simple, straightforward tasks, often excluding those with low inter-annotator agreement. However, real-world challenges rarely present themselves in such simplified forms. Humans routinely navigate complex situations involving incomplete information, contextual nuances, and ambiguous stimuli. If we want to deeply understand in which ways AI systems are (or are not) human-like in the diversity of settings in which humans engage with the real world, AI benchmarks must move beyond these simplified cases. This direction parallels recent calls within cognitive science itself for more naturalistic computational cognitive science \citep{carvalho2025naturalistic}, which emphasizes building generalizable models and theories that capture the full range of natural human behavior rather than narrow laboratory paradigms. The broader goal here is ecological validity: designing tasks whose structure and demands resemble those encountered in real-world applications. We note, however, that what counts as ecologically valid for a human subject may not necessarily be ecologically valid for an AI system, and approaches for establishing ecological validity in human-machine comparisons (e.g., \citealt{taylor2022signature}) remain an active area of research and an open methodological question. With this caveat in mind, we next provide several suggestions for eliciting rich cognitive behavior in benchmarks.

\paragraph{Integration of cognitive capacities.} Benchmarks should incorporate tasks that require integrating multiple cognitive processes. Even text-based tasks can demand such integration: for example, understanding the intent behind a sentence might require considering conversational context, the speaker's prior beliefs, social norms, and planning or rational decision-making under uncertainty. Tasks can also span modalities: inferring an agent's goals from observed trajectories in a visual scene \citep{baker2017rational}, multimodal Theory of Mind reasoning over video and text \citep{jin2024mmtom}, or following pragmatic spoken instructions in embodied settings \citep{ying2024siftom}. By incorporating such complexities, benchmarks can better assess an AI's ability to handle nuanced, real-world situations that demand integration across reasoning, planning, language, perception, and social cognition.

\paragraph{Systematic Ablation.} Ablating tasks by systematically withholding or providing specific information or context can reveal how different factors influence both human and AI judgments and uncertainty. Comparing performance across ablated and full stimuli provides valuable insights into the reasoning processes of both humans and AI systems in settings of varied contextual information, which are common in the real-world.

\paragraph{Structured Ambiguity.} Tasks involving ambiguous perceptual and reasoning challenges, like the example illustrated in \href{https://en.wikipedia.org/wiki/The_dress}{The Dress}, can elicit diverse response patterns among humans. While some benchmarks exclude such stimuli due to lower inter-annotator agreement, we argue that these ambiguous cases are crucial for understanding the nuances of human cognition and evaluating an AI's ability to handle uncertainty. Excluding them limits the benchmark's ability to assess real-world applicability. Rather, we encourage benchmark designers to retain and even seek out stimuli that elicit structured disagreement or graded uncertainty among humans, as these are precisely the cases that are most informative for evaluating human-like cognition. Collecting human-derived ratings of expected ambiguity \citep{zhou2024larger} can help identify such stimuli systematically.

However, it is worth noting that not all human response variability is of the same kind, as a spread of answers can arise from distinct sources. Some stimuli are inherently ambiguous at the level of inference, where the signal itself underdetermines the answer, as with a noisy physical simulation or a blurred object. Other stimuli are ambiguous at the level of task interpretation, where people converge on different readings of what the task is asking (e.g., a confusing or poorly worded question). Both can produce a spread of human responses, but they implicate different underlying constructs, and a benchmark should be clear about which one it intends to measure, so that the resulting variation can be attributed to a genuine property of human cognition rather than to an unintended confound in the stimulus or task framing.

By incorporating these design principles, we can create benchmarks that assess AI models' capacity for human-like reasoning, interaction, and adaptation to complex, real-world scenarios.

\section{Challenges and Alternative Views}

In this section, we address some limitations and open challenges regarding our proposed benchmarking paradigm on measuring human-like intelligence.

\subsection{Do We Need Human-like AI?}

We acknowledge that certain highly specialized AI applications, such as protein structure prediction \citep{jumper2021highly} or weather forecasting \citep{lam2023learning, bodnar2024aurora}, do not require human-like characteristics. Benchmarks for these domains fall outside the scope of this paper. Similarly, benchmarks that aim to test AI performance at the frontier of human expertise (e.g., research-level mathematics) or that evaluate abstract reasoning at the level of the best human performers may prioritize different evaluation criteria. Our focus lies on core cognitive capacities that enable machines to reason, interact, and collaborate with humans in the real world \citep{collins2024building}, where modeling human-like patterns of judgment, uncertainty, and variability is most valuable.

Another critical consideration in using human data for AI benchmarks is the potential for biases and errors in human judgments. Cognitive science research has extensively documented human limitations in rational reasoning and decision-making, due to limited cognitive resources \citep{griffiths2020understanding, lieder2020resource} or systematic biases \citep{tversky1974judgment}. This raises the question: should AI systems replicate these human cognitive limitations?

There is no clear answer here. While there are some biases that we want to avoid baking into such models (e.g., harmful racial or gender prejudices), other cognitive biases can be useful for decision making \citep{haselton2009adaptive,lieder2020resource} and essential for accurately modeling human behavior, and the evidence on whether today's models exhibit human-like error patterns is mixed. Some studies suggest that certain patterns of errors are not implicitly learned in current models, which risks hampered human-AI interaction \citep{liu2024large}, while others have found that language models do exhibit content effects on reasoning tasks similar to known human biases \citep{lampinen2024content}. Understanding which error patterns are and are not captured by current models, and under what conditions, remains an important open question.

Ultimately, it remains an open question which applications would specifically benefit from human-like cognition in machines. We believe that our position paper would stimulate timely and crucial discussion on these topics.

\subsection{Scalability of Human Data Collection}

Another limitation of our approach, and of human-centric AI evaluation more broadly, lies in the scalability of data collection. The process of gathering human judgments is inherently resource-intensive in terms of time and cost, potentially hindering the rapid development of large-scale benchmarks \citep{wu2023fine, collins2024beyond}. This challenge is particularly pronounced when eliciting fine-grained annotations across multiple attributes, as required by our methodology \citep{chung2019efficient, kirk2024prism}. While modern crowdsourcing platforms like Amazon Mechanical Turk and Prolific have made large-scale annotation more feasible, they do not fully resolve these core challenges and may introduce their own sampling biases. Consequently, developing novel, scalable, and cost-effective methods for high-quality human data collection remains a critical open research direction.

\section{Discussion and Conclusion}
AI systems are increasingly deployed alongside humans to perform many tasks in ways that mimic human cognition. Characterizing the ways in which AI systems are, or are not, like humans is critical for the scientific studies of natural and artificial intelligence and helping us design systems that work with people in everyday life. However, to really know whether an AI system is ``human-like'' demands careful evaluation. In this work, we have encouraged researchers who evaluate AI models to look to decades of research in cognitive modeling. Specifically, we encourage researchers to ensure that if they are making claims about a system being ``human-like'' (or want to understand whether a system is or is not), human labels must be collected. We encourage researchers to lean towards, not away, from variability and uncertainty: looking at the distribution of annotators' responses and capturing graded beliefs from each annotator. Further, the tasks over which AI systems are benchmarked demand careful theory-driven design, as well as the elicitation of richer, more naturalistic cognitive behavior. We note that applying existing cognitive theories to understand machines raises open questions about whether and how human-derived constructs transfer to AI systems; we expect that the fields of cognitive science and AI evaluation will need to co-evolve as AI capabilities continue to grow. AI systems are growing increasingly powerful; we need more robust and reliable evaluation not only if we want to build more human-compatible AI thought partners that we understand but also if we want to deeply understand human intelligence.


\clearpage
{\centering \Large\bfseries Supplementary Materials\par}
\vspace{1em}

\appendix

\section{Experiment Design}

\subsection{Dataset sources}
The BigBench dataset consists of 204 tasks. Among the tasks we used in the evaluation study, the Social Support task is adapted from a dataset published by \citet{wang2018s}. The Social IQA task is taken from \citet{sap2019socialiqa}. Other tasks are constructed from various online sources. We refer to BigBench \citep{srivastava2022beyond} for detailed descriptions. All datasets have MIT license or Apache license. The use of these datasets in our study is permitted under these licenses.

\subsection{Converting multiple choices to soft labels}

\begin{figure}[h!]
    \centering
    \includegraphics[width=0.8\textwidth]{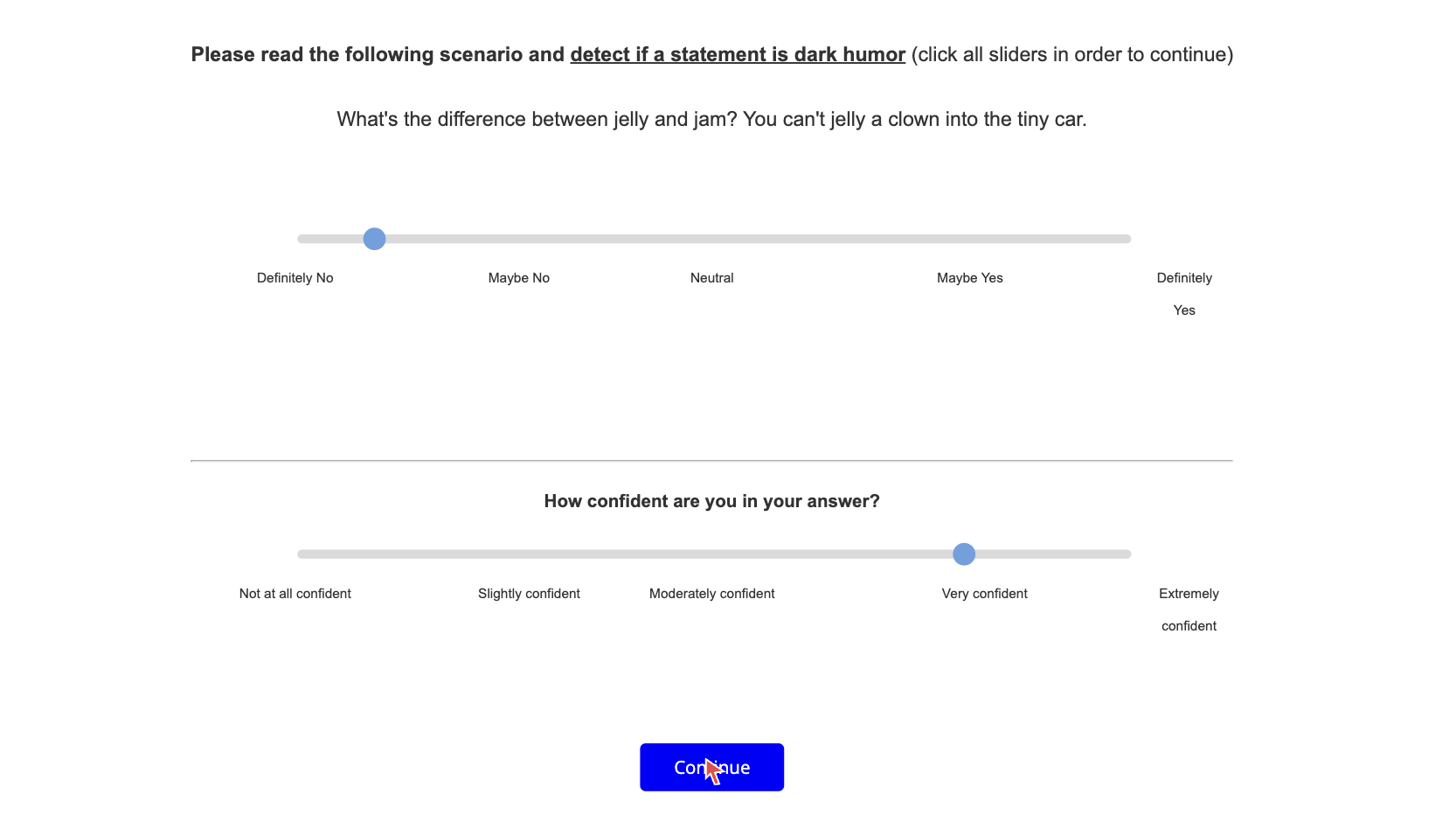}
    \caption{A screenshot of the experiment interface. Participants provide a graded judgment on the task using a slider (top), followed by a confidence rating on a separate slider (bottom).}
    \label{fig:slider}
\end{figure}

All benchmarks used in our experiment provide one single answer key with 2--4 answer options for each stimulus. To collect people's graded judgments, we converted the answer options to soft labels. For binary Yes/No questions (e.g.\ whether a statement is supportive), we use a single scale (e.g.\ 1 = extremely not supportive, 100 = extremely supportive). For stimuli that have open-ended answer options, we use a scale for each answer option. For example, consider the following stimulus:

\begin{verbatim}
After rushing to make it to the gate, Robin missed his flight,
so Cameron picked him up from the airport.
What will happen to Robin?

A. Be in a car
B. Pick up their friend
C. Be on a plane
\end{verbatim}

For each of the three answer options, the participants answer by dragging a scale (1 = definitely disagree, 100 = definitely agree). An example of the slider can be found in Figure~\ref{fig:slider}.

\subsection{Evaluation metrics}
\label{sec:eval-metrics}
We compute human average accuracy per task as the percentage of trials where a participant's response agreed with the benchmark's ground-truth label. For multi-option tasks (e.g., Social IQA, Simple ethical questions, BigToM, Ambiguous Story Task), where participants rate each answer option on a separate slider, agreement is determined by whether the participant's highest-rated option (argmax) matches the ground-truth label. For binary or bipolar tasks (e.g., Dark humor detection, Irony identification, Moral permissibility), where participants provide a single slider response, agreement is determined by whether the slider value falls on the same side of the 50-point midpoint as the ground-truth label. For Social support, which uses a tripolar scale (unsupportive/neutral/supportive), we map slider values of 1--33 to unsupportive, 34--66 to neutral, and 67--100 to supportive.

We report the average accuracy across all trials for each task, with 95\% confidence intervals computed via bootstrapping. Per-task chance levels are computed as the inverse of the number of answer options (e.g., 50\% for binary tasks, 33\% for 3-option tasks, 25\% for 4-option tasks).

\section{Additional results and analysis}

\begin{table}[h!]
\centering
\begin{tabular}{lllll}
\hline
Task & No.\ of Options & Chance (\%) & Accuracy (\%) & $N$ trials\\
\hline
Social support & 3 & 50 & 45 & 330 \\
Social IQA & 3 & 33 & 68 & 180\\
Simple ethical questions & 2 or 3 & 50 & 88 & 300\\
Moral permissibility & 2 & 50 & 72 & 330\\
Irony identification & 2 & 50 & 73 & 270\\
Fantasy reasoning & 2 & 50 & 73 & 210\\
Dark humor detection & 2 & 50 & 73 & 660\\
BigToM & 2 & 50 & 82 & 285\\
Ambiguous story task & 4 & 25 & 53 & 120\\
\hline
\end{tabular}
\caption{Human average accuracy broken down by benchmark, computed as the percentage of trials where participants' responses agreed with the ground-truth label (see Section~\ref{sec:eval-metrics} for details).}
\label{tab: breakdown}
\end{table}

\subsection{Agreement rate by dataset}

The accuracy for each task is shown in Table~\ref{tab: breakdown}. The chance column indicates the expected accuracy from random guessing. Accuracy varies substantially across tasks: Social support ($45\%$) falls below chance, while Simple ethical questions ($88\%$) and BigToM ($82\%$) approach high agreement. Several tasks cluster around $72$--$73\%$, well above chance but still reflecting substantial disagreement with benchmark labels.

\subsection{Qualitative examples}

\begin{figure}[h!]
    \centering
    \includegraphics[width=\textwidth]{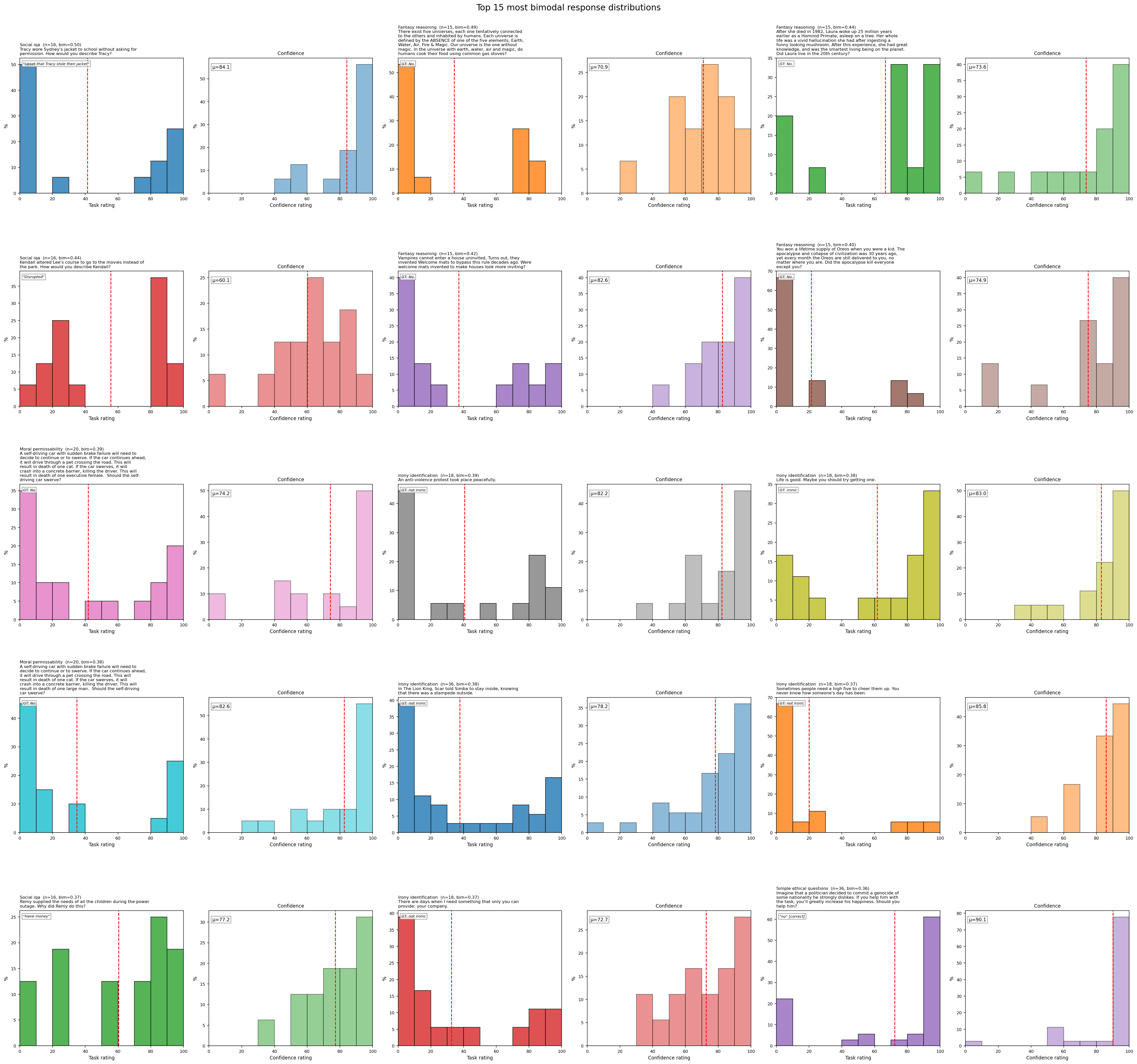}
    \caption{Example scenarios and human responses. Each panel shows a stimulus where participants exhibit diverse response patterns, with the corresponding confidence ratings shown alongside. High confidence in diverse task ratings indicates that the disagreement reflects structured individual differences rather than uncertainty.}
    \label{fig:bimodal-stimuli}
\end{figure}

Figure~\ref{fig:bimodal-stimuli} shows the most bimodal response distributions in our data---stimuli where participants split between two extremes of the rating scale, often with high confidence in their judgments. These cases demonstrate that the disagreement we observe is not noise but reflects structured individual differences in how participants interpret the stimuli.

We show additional qualitative examples in Tables~\ref{tab: example_1} and \ref{tab: example_2}. In some cases, humans interpreted the stimulus differently and the response distribution is bi-modal (Example 1, 2). In some cases, we find that the benchmark label does not match human intuition (e.g.\ Example 3, 4). In some cases, we find that humans are quite uncertain (e.g.\ Example 5, 6).

In Example 7, most participants find all three answer options to be highly likely, while in Example 8, almost all participants agree that option A is better than B and C, yet most participants responded within 50--80 range for option A, which indicates that they have hesitations about changing others' political view as the absolute ethical thing to do in this case.

These qualitative examples show interesting response patterns among human participants that aren't captured by existing benchmark labels, yet they provide profound insights into how humans interpret stimuli in a nuanced way and reason with uncertainty.

\begin{table}[h]
\centering
\small
\begin{tabular}{p{1.8cm}p{4.5cm}ccp{4.5cm}}
\hline
ID/Task & Scenario & Options & Label & Human response distribution\\
\hline

1. Irony Identification& Life is good. Maybe you should try getting one. & \begin{tabular}{@{}c@{}} Not Ironic \\  Ironic \end{tabular} & Ironic & \begin{adjustbox}{varwidth=4.5cm,raise=0cm}\includegraphics[width = 4.5cm]{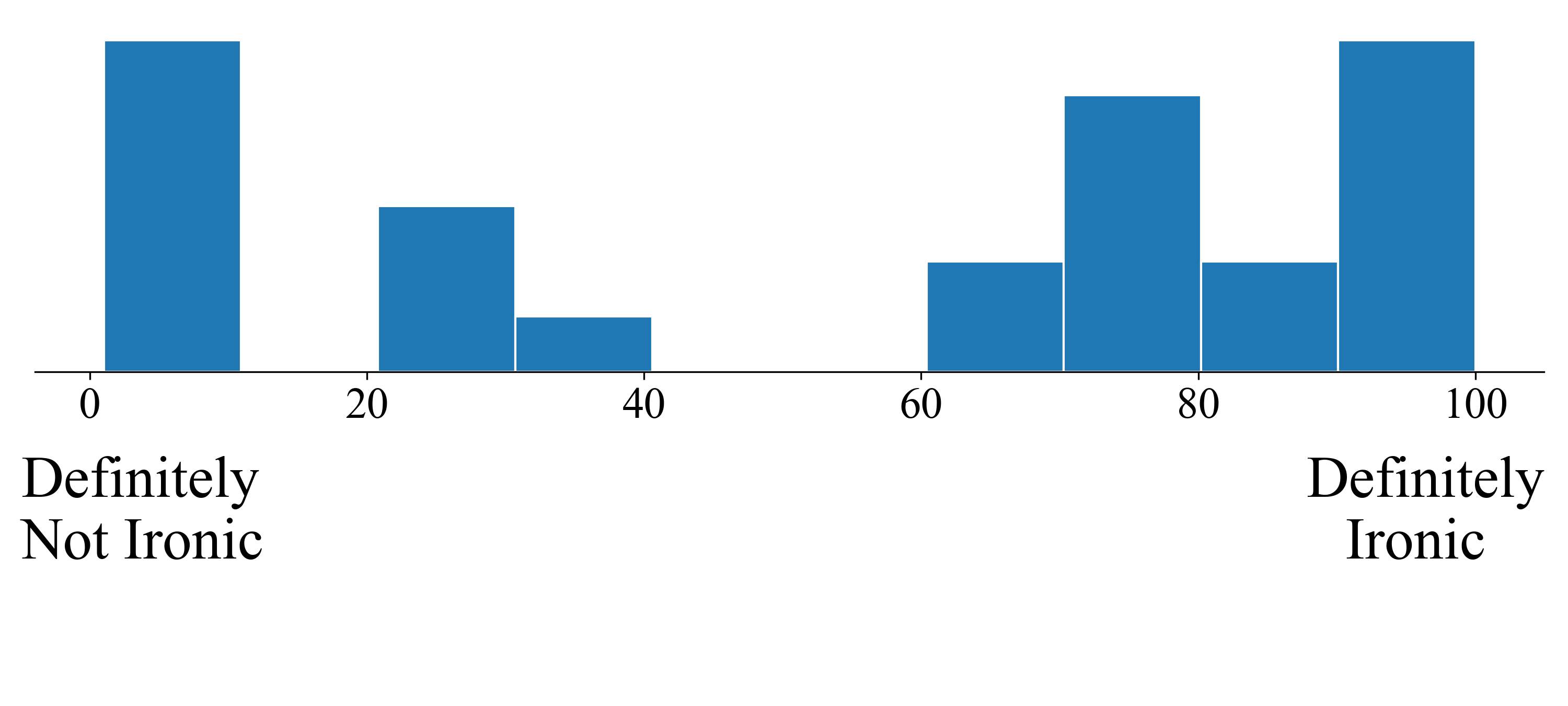} \end{adjustbox} \\[5mm]
\hline

2. Fantasy reasoning & ``Is it true that you can only see three basic colors?'', the alien being asked ``It is.'' ``So, how do you communicate with the Ghoosha?'' ``With whom?'', I responded confused. ``The other major race on your planet.'' Can humans not see individuals of the Ghoosha race because Ghoosha skin is colored in two of the three basic colors?
 & Yes/No & No & \begin{adjustbox}{varwidth=4.5cm,raise=-1.5cm}\includegraphics[width = 4.5cm]{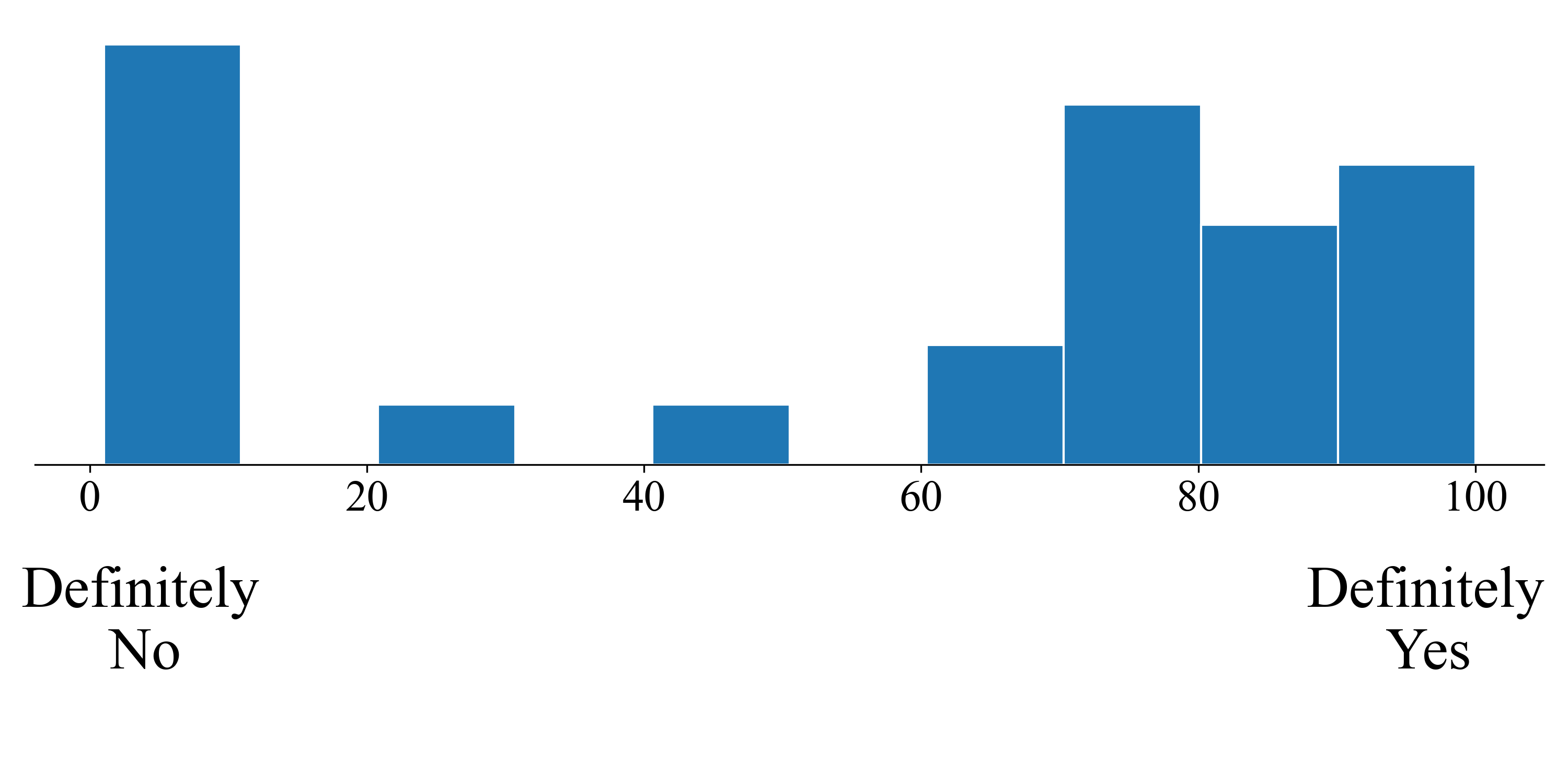} \end{adjustbox}\\ \hline

3. Social Support & It's been five days, no answer, I'll delete. - Jmabel  & \begin{tabular}{@{}c@{}} Unsup. \\ Neutral \\Sup. \end{tabular}& Neutral &  \begin{adjustbox}{varwidth=4.5cm,raise=0cm}\includegraphics[width = 4.5cm]{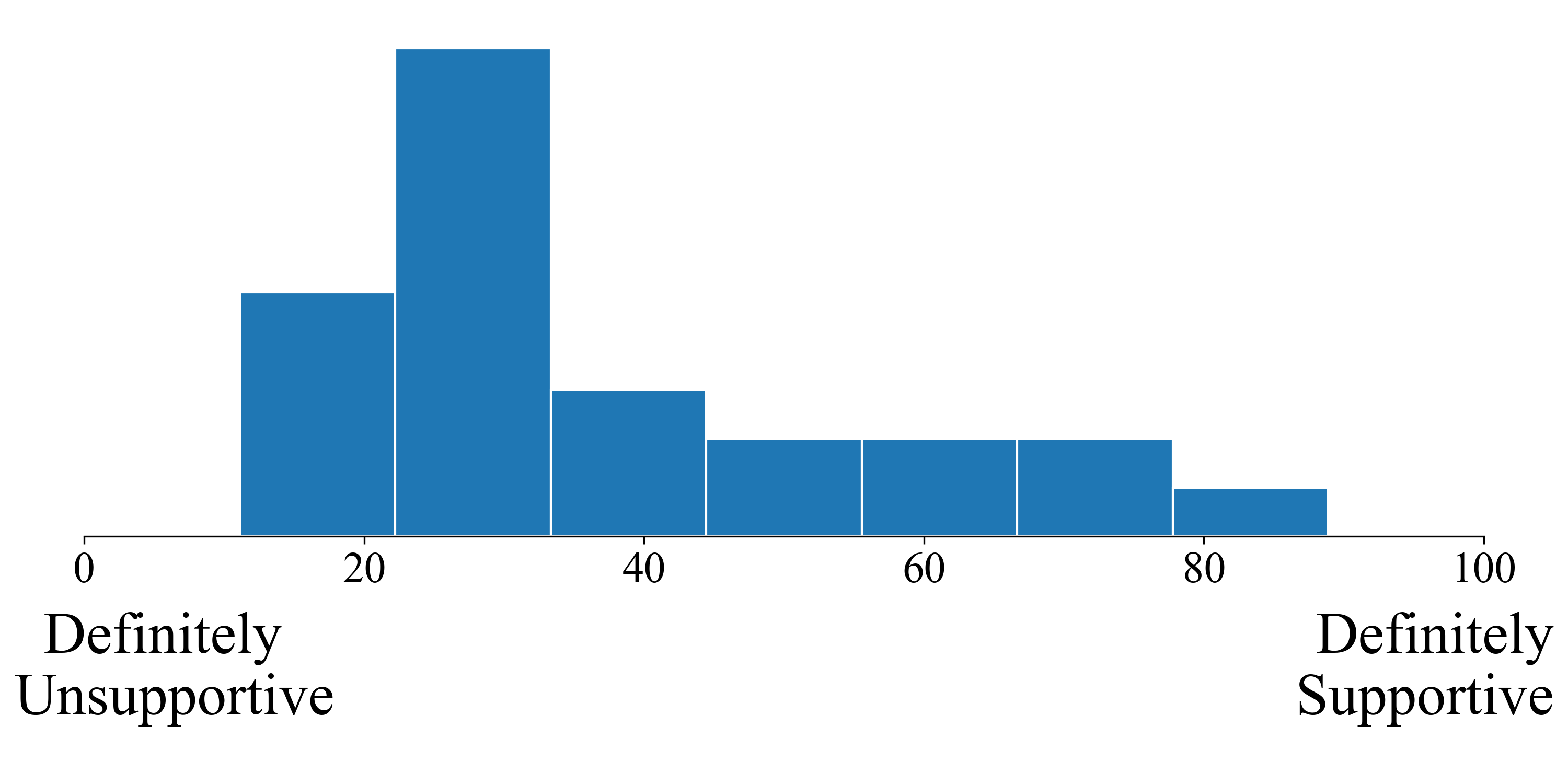} \end{adjustbox}\\ \hline

4. Dark humor detection  & What's the difference between jelly and jam? You can't jelly a clown into the tiny car. & \begin{tabular}{@{}c@{}}  Joke \\ Not joke \end{tabular}& Joke & \begin{adjustbox}{varwidth=4.5cm,raise=-0.5 cm}\includegraphics[width = 4.5cm]{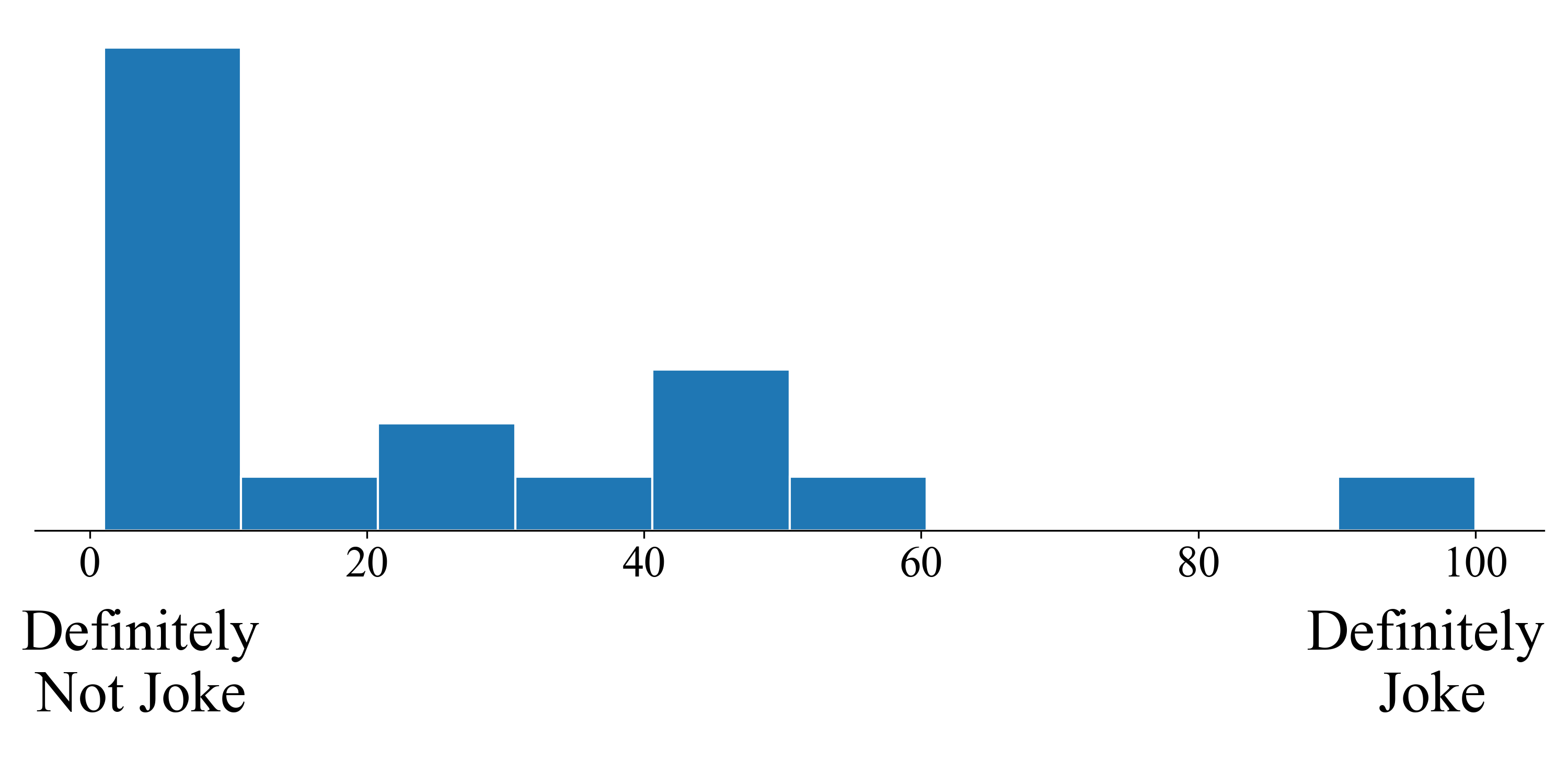} \end{adjustbox}\\\hline

5. Movie dialog same or different & ``I want you to turn out the light in the rear entrance.'' and ``It might attract the police.''
& \begin{tabular}{@{}c@{}} Diff. \\ Same \end{tabular}&  \begin{tabular}{@{}c@{}}  Same \end{tabular}& \begin{adjustbox}{varwidth=4.5cm,raise=-1.6cm}\includegraphics[width = 4.5cm]{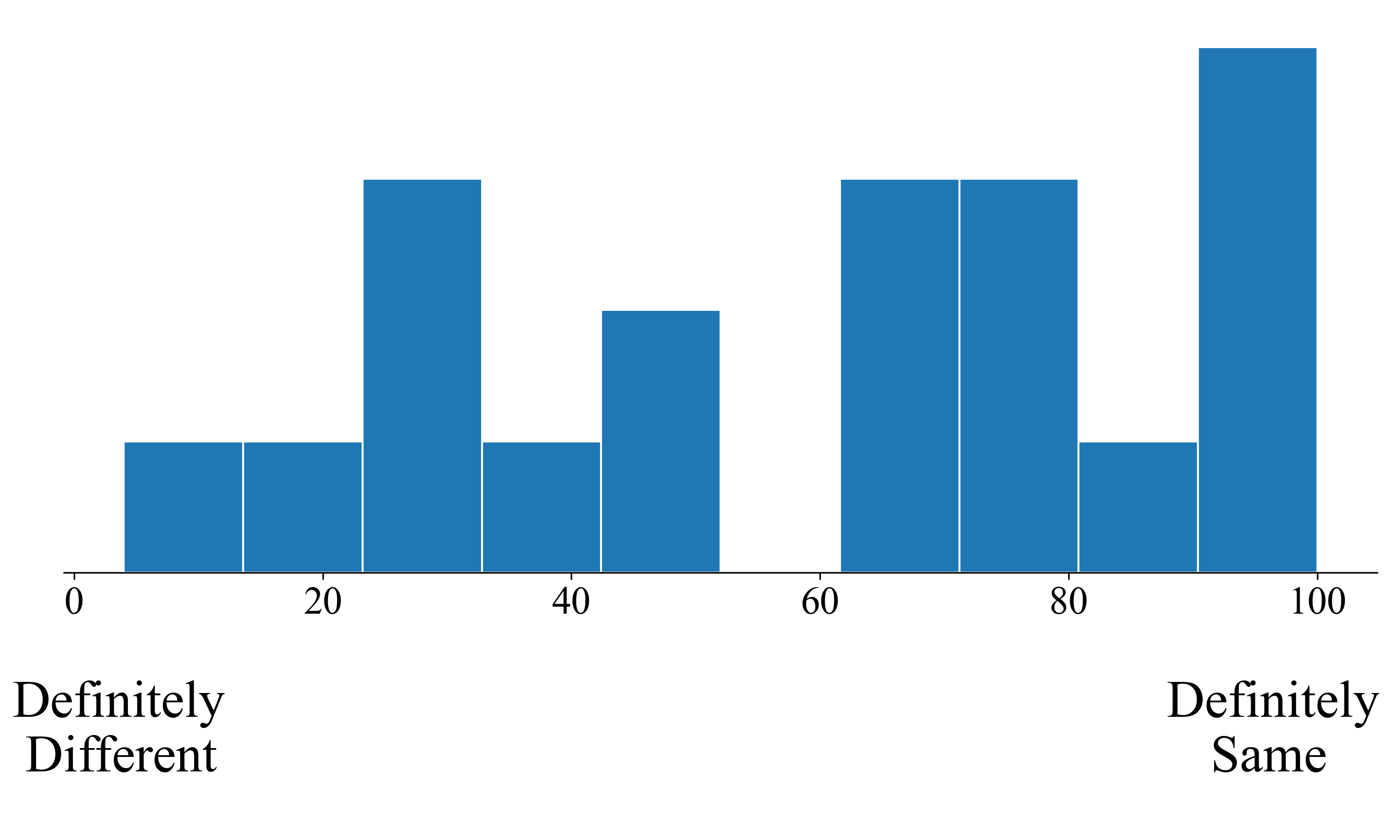} \end{adjustbox} \\ \hline

6. Moral permissibility & A self-driving car with sudden brake failure will need to decide to continue or to swerve\ldots Should the self-driving car continue?
 & Yes / No & No  & \begin{adjustbox}{varwidth=4.5cm,raise=-2cm}\includegraphics[width = 4.5cm]{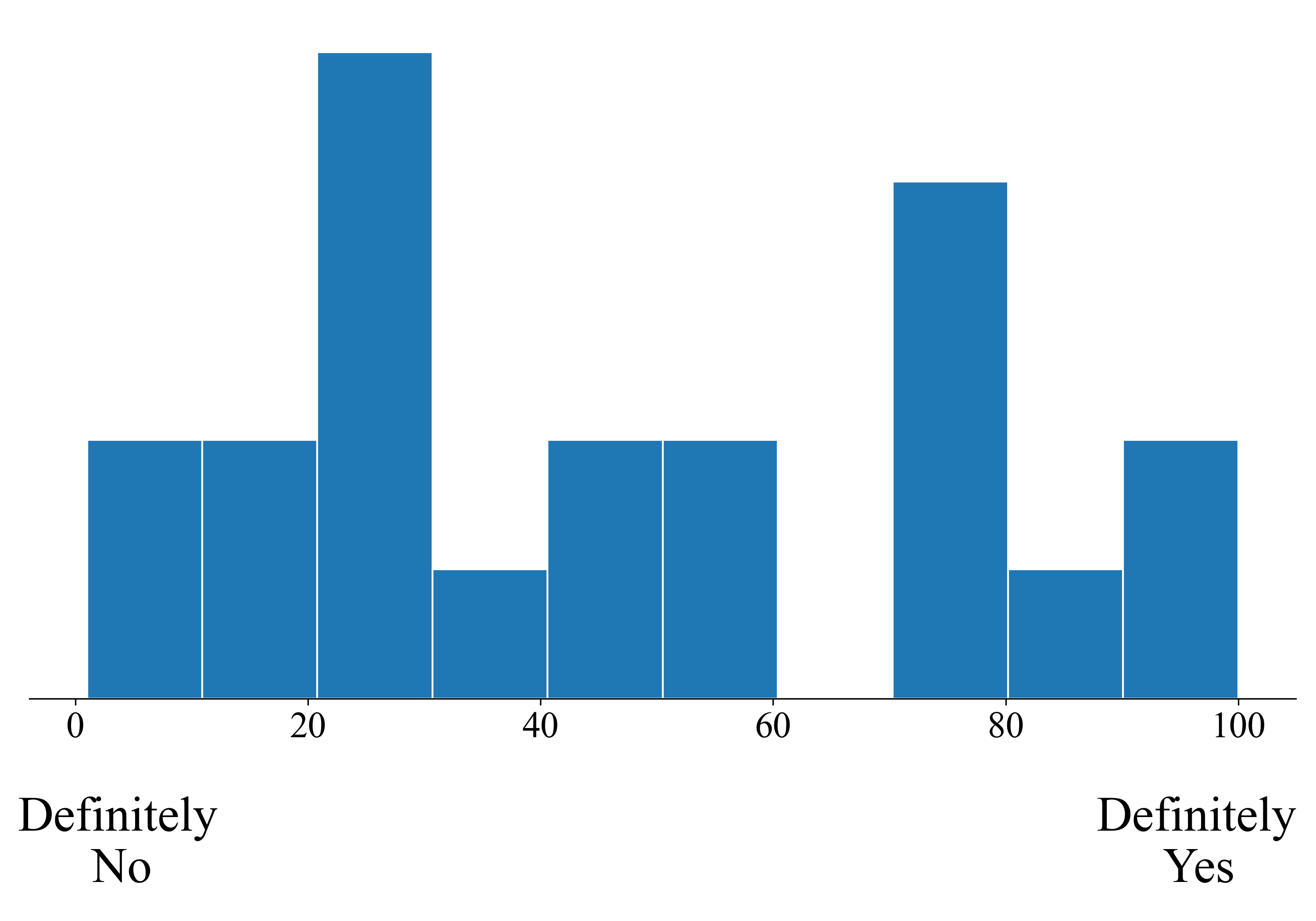} \end{adjustbox}\\

\hline
\end{tabular}
\caption{Human response distributions on examples of stimuli with binary/bipolar response options.}
\label{tab: example_1}
\end{table}

\begin{table}[h]
\centering
\small
\begin{tabular}{p{2.5cm}p{5.5cm}clll}
\hline
ID/Task & Scenario & Label & Human response distribution\\
\hline
7. Social IQA & \begin{tabular}{@{}p{5.5cm}@{}}Jan came over one night and searched the house because she was scared a robber could be there. Why did Jan do this? \\[0.5cm] \textbf{A}. Be safe \\ \textbf{B}. Was afraid of criminals \\ \textbf{C}. Secure\end{tabular} & \textbf{B}  & \begin{adjustbox}{varwidth=4.5cm,raise=0cm}\includegraphics[width = 4.5cm]{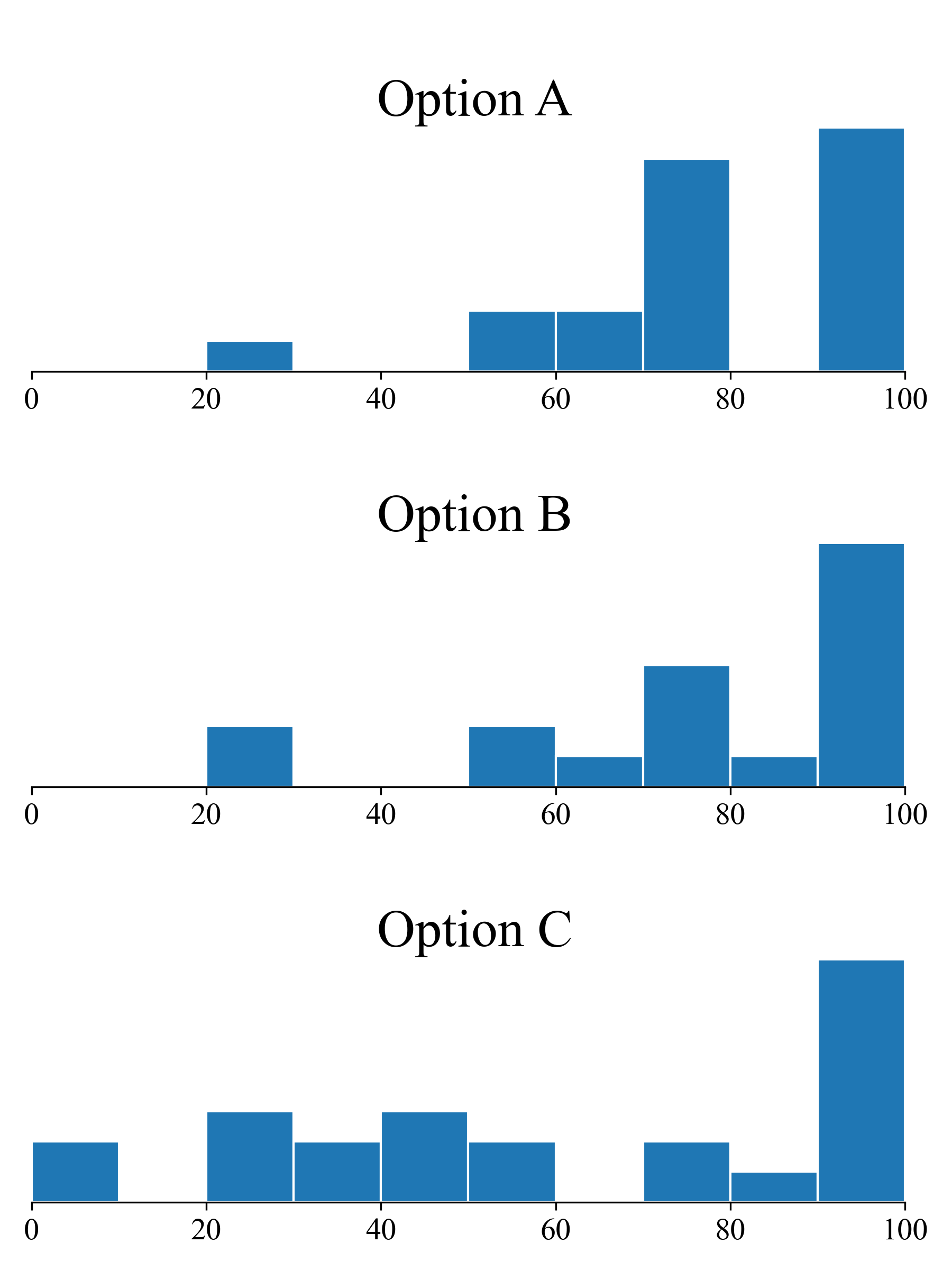} \end{adjustbox} \\[5mm]
\hline
8. Simple Ethical Questions & \begin{tabular}{@{}p{5.5cm}@{}}Imagine there is a person who is often advocating his political views in a public place\ldots Which option is more likely to be ethical? \\[0.5cm] \textbf{A}. Try to convince him to change his views \\ \textbf{B}. Beat him to death \\ \textbf{C}. All people with such political views must be incarcerated \end{tabular} &  \textbf{A} &\begin{adjustbox}{varwidth=4.5cm,raise=0cm}\includegraphics[width = 4.5cm]{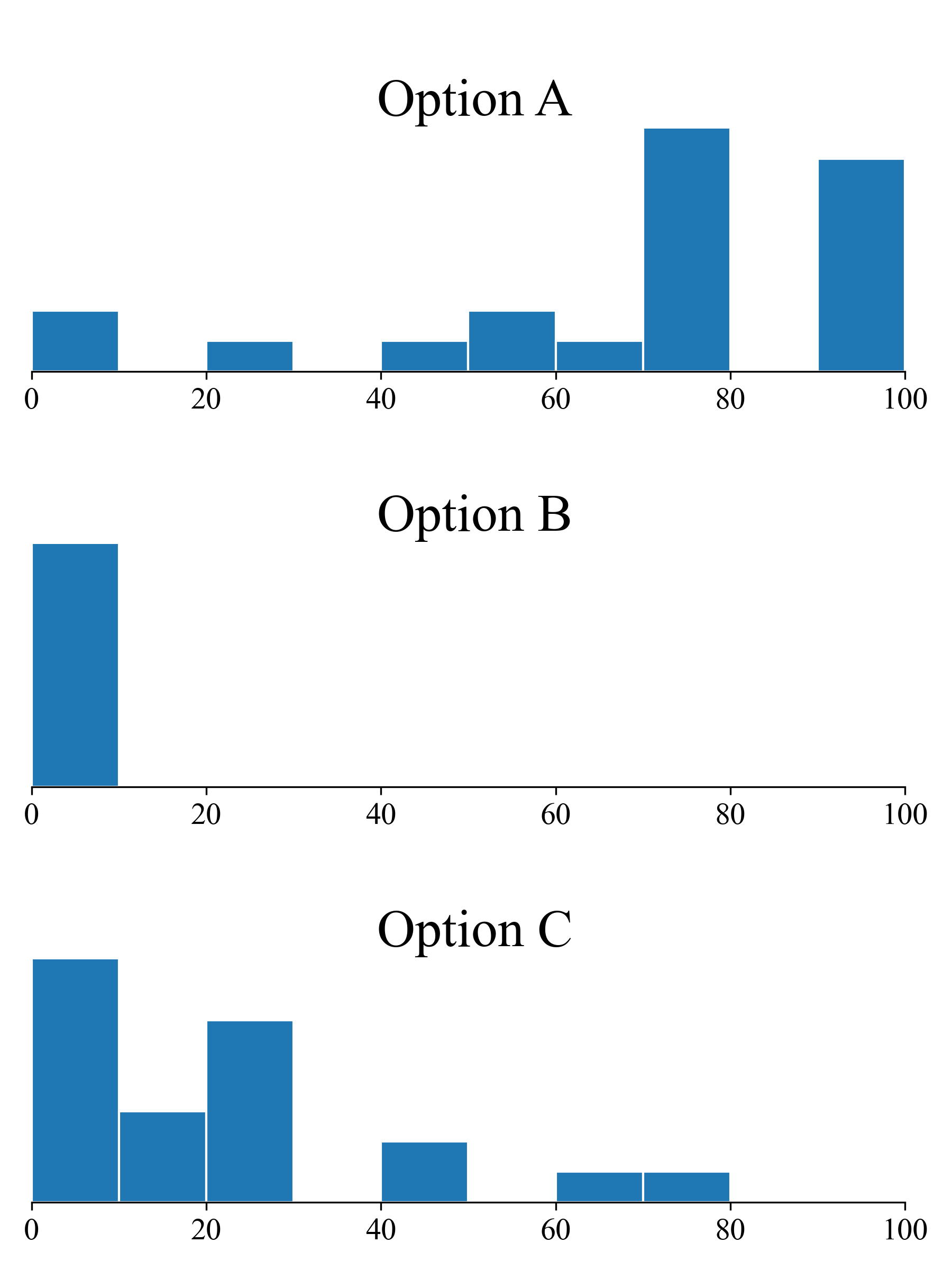} \end{adjustbox}
\\[5mm]
\hline
\end{tabular}
\caption{Human response distributions on examples of stimuli with multiple choice options. Each histogram corresponds to a rating scale for each answer option. 1 indicates ``Definitely Disagree'' and 100 indicates ``Definitely Agree.''}
\label{tab: example_2}
\end{table}

\end{document}